\documentclass[twoside]{article}

\usepackage{microtype}
\usepackage{graphicx}
\usepackage{subfigure}
\usepackage{booktabs} 
\usepackage{algorithm}

\usepackage{changes}

\usepackage{graphicx}
\usepackage{comment}
\usepackage{amsmath,amssymb} 
\usepackage{color}
\usepackage{booktabs}
\usepackage{bbm}
\usepackage[font=small,labelfont=small]{caption}

\definecolor{dark}{rgb}{1.0, 0.55, 0.0}
\usepackage{multirow}
\usepackage{hhline}
\usepackage{wrapfig}

\usepackage[algo2e,vlined,ruled]{algorithm2e}

\usepackage{url}

\usepackage{enumitem}

\def\*#1{\mathbf{#1}}
\newcommand{\GP}{$\mathcal{GP}$}
\newcommand{\GPf}{\texttt{$\mathcal{GP}$-ND}}
\def\eg{\emph{e.g.}}

\def\ie{\emph{i.e.}}

\usepackage{bbold}

\newcommand{\barx}{\bar{x}}
\newcommand{\bary}{\bar{y}}
\newcommand{\bff}{\mathbf{f}}
\newcommand{\bu}{\mathbf{u}}
\newcommand{\by}{\mathbf{y}}
\newcommand{\bx}{\mathbf{x}}

\newcommand{\bz}{\mathbf{z}}
\newcommand{\bmm}{\mathbf{m}}
\newcommand{\bk}{\mathbf{k}}

\newcommand{\EE}{\mathbb{E}}

\newcommand{\KL}{D_{\mathrm{KL}}}

\newcommand{\OO}{\mathcal{O}}

\newcommand{\sigmaobs}{\sigma_{\rm obs}}

\newcommand{\argmin}{\text{argmin}}

\usepackage[accepted]{aistats2022}
\begin{document}

\twocolumn[

\aistatstitle{Obstacle-aware Gaussian Process Regression}

\aistatsauthor{ Gaurav Shrivastava}

\aistatsaddress{University of Maryland, College Park} ]

\begin{abstract}
Obstacle-aware trajectory navigation is an important need for many systems. For instance, in real-world navigation tasks, an agent would want to avoid obstacles, like furniture items in a room when planning a trajectory to traverse. Gaussian Process (\GP) regression in its current construct is designed to fit a curve on a set of datapairs, with each pair consisting of an input point `$\bx$' and its corresponding target regression value `$y(\bx)$' (a \textit{positive} datapair). But, in order to account for obstacles, for an input point `$\bar{\bx}$', we would want to constrain the \GP\ to avoid a target regression value `$\bar{y}(\bar{\bx})$' (a \textit{negative} datapair).
Our proposed approach, `\GPf' or Gaussian Process with Negative Datapairs, fits over the anchoring positive datapairs while avoiding the negative datapairs. Specifically, we model the negative datapairs using small blobs of Gaussian distribution and maximize its KL divergence from the \GP. Our framework jointly optimizes for both the positive and negative datapairs. Our experiments demonstrate that \GPf\ performs better than the traditional \GP\ learning. Furthermore, our framework does not affect the scalability of Gaussian Process regression and helps the model converge faster as the size of the data increases. 
\end{abstract}

\section{Introduction}
\begin{figure*}[t]
\vspace{-0.05in}
	\centering
    \includegraphics[width=0.8\linewidth]{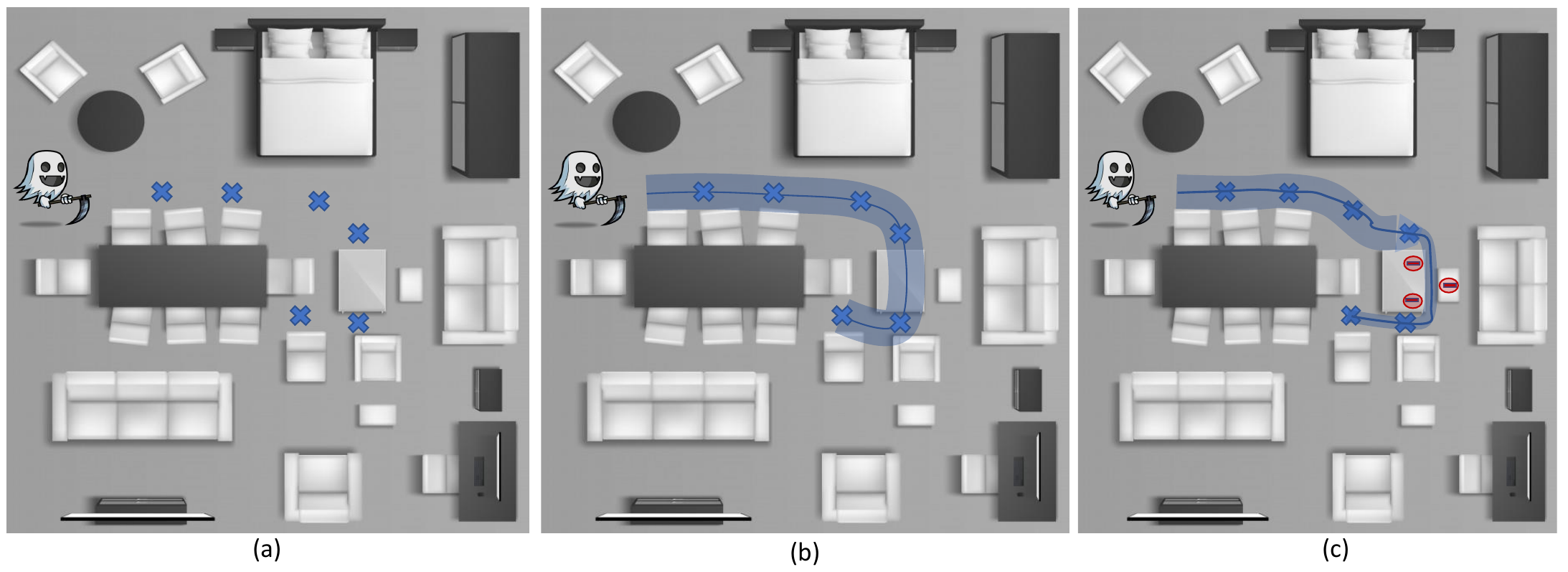}
    \caption{An illustration of our problem setup. (a) top view of the room where the agent wants to travel through particular locations while avoiding obstacles; (b) the agent has been given the location of the positive datapairs that are needed to be covered in its trajectory. Since the number of these observed points is low, the agent is not able to avoid the obstruction (coffee table) while forecasting its course; (c) the agent is given both the positive datapairs which it needs to reach along with negative datapairs (area of influence is given by shaded red region) that should be avoided during the modeling of future trajectory.}
    \label{teaser}
\end{figure*}
Gaussian processes are one of the most studied model classes for data-driven learning as these are nonparametric, flexible function classes that requires little prior knowledge of the process. Traditionally, GPs have found their applications in various fields of research, including Navigation systems (\eg, in Wiener and Kalman filters)~\cite{jazwinski2007stochastic}, Geostatistics, Meteorology~\cite{handcock1993bayesian} and Machine learning~\cite{rasmussen2003gaussian}. The wide range of applications can be attributed to the property of \GP s to model the target uncertainty by providing the predictive variance over the target variable. 
Gaussian process regression in its current construct anchors only on a set of \textit{positive} datapairs, with each pair consisting of an input point and its desired target regression value. However, in some cases, more information is available in the form of datapairs, where at a particular input point, we want to avoid a range of regression values during the curve fitting of \GP. We designate such data as \textit{negative} datapairs. 

An illustration where modeling such negative datapairs would be extremely beneficial is given in Fig.~\ref{teaser}. In Fig.~\ref{teaser}(b), an agent wants to model a trajectory such that it covers all the positive datapairs marked by ``$\mathbf{x}$''. However, it is essential to note that the agent would run into an obstacle if it models its trajectory based only on the positive datapairs. We can handle this problem of navigating in the presence of obstacles in two ways. One way is to get a high density of positive datapairs near the obstacle, and the other more straightforward approach is to just mark the obstacle as a negative datapair. The former approach would unnecessarily increase the number of positive datapairs for \GP\ to regress, hence, it may run into scalability issues. However, in the latter approach, if the obstacle is denoted as a negative datapair with a sphere of negative influence around it which needs to be avoided (Fig.~\ref{teaser}.c), the new trajectory can be modeled with fewer datapairs that account for all obstacles in the way. Various \GP\ methods in their current framework lack the ability to incorporate these negative datapairs for the regression paradigm.

\textbf{Contributions}: In this paper, we explore the concept of negative datapairs. We provide a simple yet effective \GP\ regression framework, called \GPf\, which anchors on the positive datapairs while avoiding the negative datapairs. Specifically, our key idea is to model the negative datapairs using a small Gaussian blob and maximize its KL divergence from the \GP\ . Our framework can be easily incorporated for various types of \GP\ models (\eg, exact, SVGP~\cite{hensman2013gaussian}, PPGPR~\cite{jankowiak2019parametric}) and works well in the scalable settings too. We empirically show in $\S$\ref{sec:expts} that the inclusion of negative datapairs in training helps with both the increase in accuracy and the convergence rate of the algorithm.

\begin{figure*}[t]
    \centering
    \vspace{-0.1in}
    \includegraphics[width=0.30\linewidth]{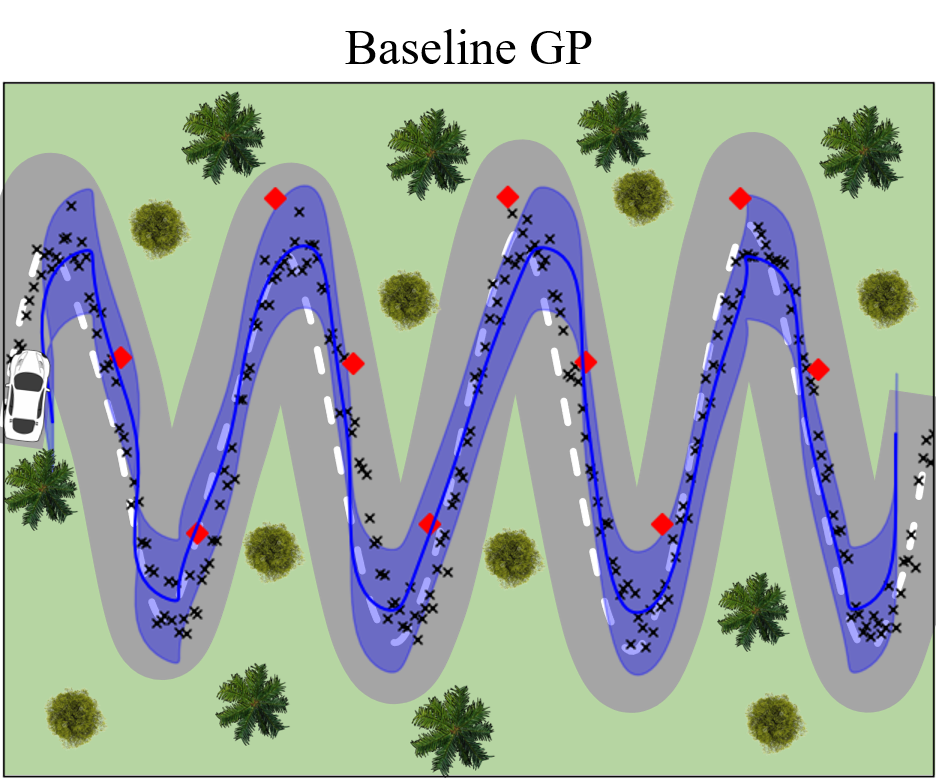}
    \hfill
    \includegraphics[width=0.304\linewidth]{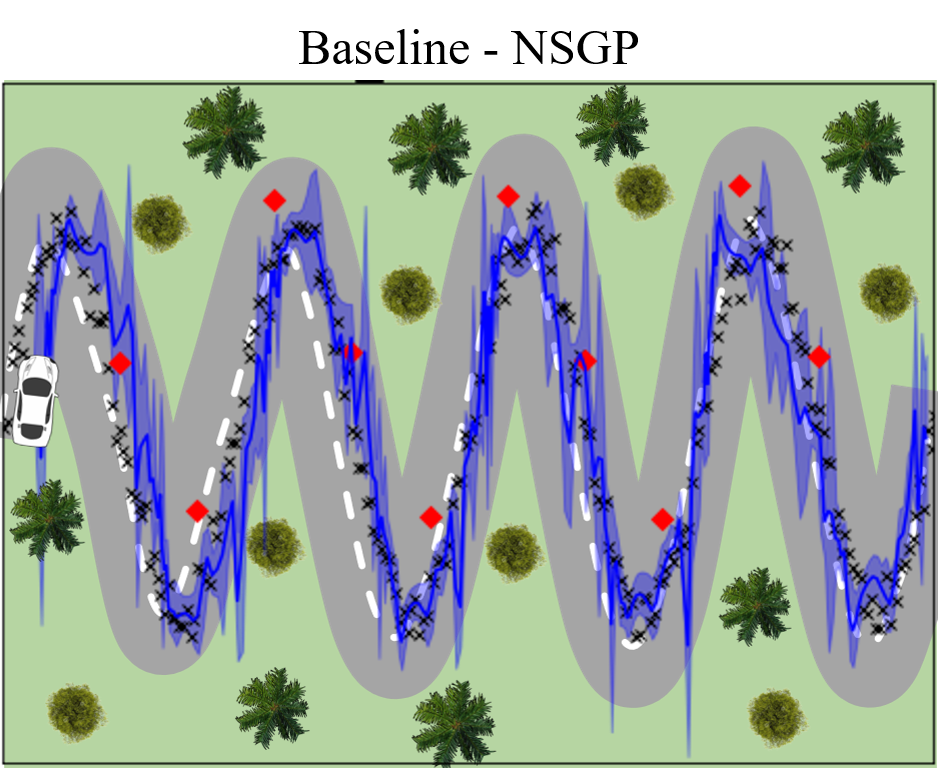}
    \hfill
    \includegraphics[width=0.30\linewidth]{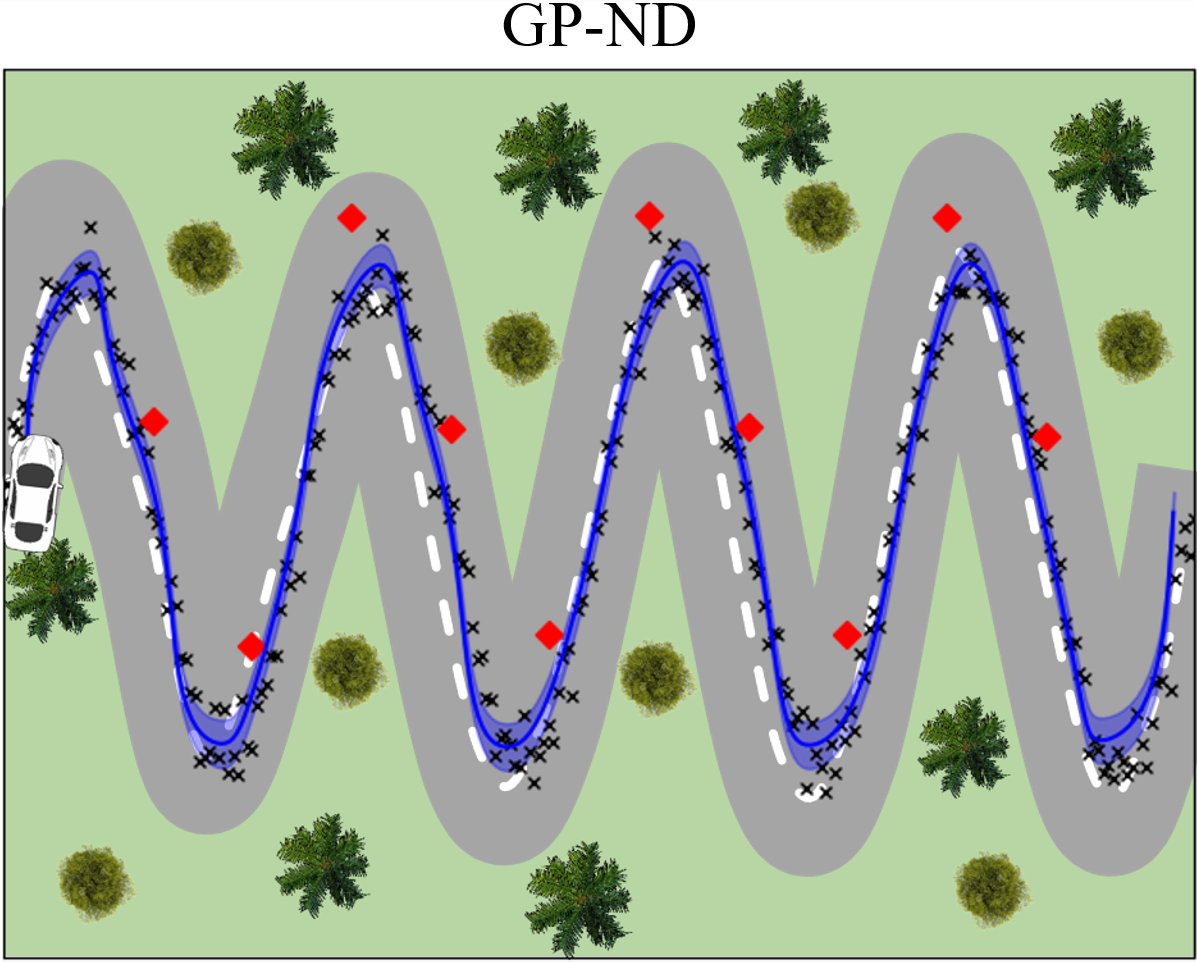}\\
    \vspace{0.1in}
    \includegraphics[width=0.6\linewidth]{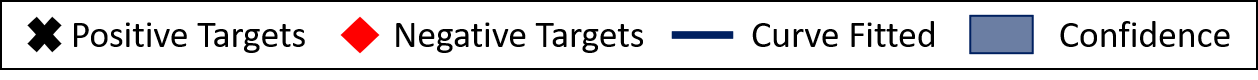}\\
    \caption{Trajectory prediction with \GPf\ regression framework: The figures compare trajectory prediction in a 2D-virtual environment using the (a) baseline \GP\  and (b) NS\GP\ framework  vs. (c) the \GPf\ framework. The car is navigating through the forest and our aim is to avoid the roadblocks marked in {`\color{red}red} while maintaining the car's proximity to the `\textbf{black}' trajectory markers. The baseline \GP (or classical \GP ) framework only uses the positive datapairs whereas NS\GP\ and our proposed \GPf\ framework uses both the positive \& negative datapairs for prediction of agent's trajectory. (b) depicts the NS\GP\ framework with the hyperparameter $\lambda = -1$ (c) depicts the \GPf\ framework with the hyperparameter $\beta = 3$, $\sigma = 0.1$}
    \label{fig:vir_scene}
    \vspace{-0.1in}
\end{figure*}
\section{Review of \GP\ Regression}\label{sec:review}
We briefly review the basics of Gaussian Process regression, following the notations in \cite{wilson2015thoughts}. For more comprehensive discussion of \GP s, refer to \cite{rasmussen2003gaussian}.

A Gaussian process is a collection of random variables, any finite number of which have a joint Gaussian distribution \cite{rasmussen2003gaussian}. We consider a dataset $\mathcal{D}$ with $n$ D-dimensional input vectors, $X=\{\bx_1, \cdots, \bx_n\}$ and corresponding $n\times 1$ vector of targets $\mathbf{y} = (y(\mathbf{x}_1), \cdots, y(\mathbf{x}_n))^T$. The goal of \GP\ regression is to learn a function $f$ that maps elements from input space to a target space, \ie, $y(\mathbf{x}) = f(\mathbf{x})+\epsilon$ where $\epsilon$ is i.i.d. noise.
If $f(\mathbf{x})\sim \mathcal{GP}(\mu, k_\theta)$, then any collection of function values $\bff$ has a joint multivariate normal distribution given by,
\vspace{-0.05in}
\begin{equation*}\label{eqn:useless}
\bff = f(X) = [f(\mathbf{x}_1), \cdots, f(\mathbf{x}_n)]^T\sim \mathcal{N}\left(\mathbf{\mu}_X, K_{X, X}\right)
\vspace{-0.05in}
\end{equation*}
with the mean vector and covariance matrix defined by the functions of the Gaussian Process, as $(\mathbf{\mu}_X)_i=\mu(x_i)$ and $(K_{X, X})_{ij} = k_\theta(\mathbf{x}_i, \mathbf{x}_j)$. The kernel function $k_\theta$ of the \GP\ is parameterized by $\theta$. Assuming additive Gaussian noise, $y(\mathbf{x})|f(\mathbf{x})\sim\mathcal{N}(y(\mathbf{x}); f(\mathbf{x}), \sigma^2)$, then the predictive distribution of the \GP\ evaluated at the $n_*$ test points indexed by $X_*$, is given by
\begin{equation}\label{eqn:mean-cov}
\begin{split}
    &\bff_*|X_*, X, \by, \theta, \sigma^2 \sim \mathcal{N}\left(E[\bff_*], \text{cov}(\bff_{*})\right),\\
    & E[\bff_*] = \mu_{X_{*}} + K_{X_*, X}\left[K_{X, X} + \sigma^2I\right]^{-1}\by,\\
    & \text{cov}(\bff_*) = K_{X_*, X_*} - K_{X_*, X}\left[K_{X, X}+\sigma^2I\right]^{-1}K_{X, X_*}
\end{split}
\end{equation}
$K_{X_*, X}$ represents the $n_*\times n$ covariance matrix between the \GP\ evaluated at $X_*$ and $X$. Other covariance matrices follow similar conventions. $\mu_{X_*}$ is the mean vector of size $n_* \times 1$ for the test points and $K_{X, X}$ is the $n\times n$ covariance matrix calculated using the training inputs $X$. The underlying hyperparameter $\theta$ implicitly affects all the covariance matrices under consideration.
\subsection{\GP s: Learning and Model Selection}
We can view the \GP\ in terms of fitting a joint probability distribution as, 
\vspace{-0.025in}
\begin{equation}
\label{eqn:unireg}
p\left(\by, \bff | X\right) = p\left(\by|\bff, \sigma^2\right) p\left(\bff | X\right)
\vspace{-0.025in}
\end{equation}
and we can derive the marginal likelihood of the targets $\by$ as a function of kernel parameters alone for the \GP\ by integrating out the functions $\bff$ in the joint distribution of Eq.~(\ref{eqn:unireg}). A nice property of the \GP\ is that this marginal likelihood has an analytical form given by,

\vspace{-0.15in}
\begin{flalign}
\begin{split}
\mathcal{L}(\theta)&= \by^T\left(K_\theta+ \sigma^2I\right)^{-1}\by\\
&+-\frac{1}{2}\left(\log\left(\left|K_\theta+ \sigma^2I\right|\right) + N\log\left(2\pi\right)\right)
\end{split}
\label{eqn:marglikelihood}
\vspace{-0.05in}
\end{flalign}
where we have used $K_\theta$ as a shorthand for $K_{X, X}$ given $\theta$. The process of kernel learning is that of optimizing Eq.~(\ref{eqn:marglikelihood}) w.r.t. $\theta$.

The first term on the right hand side in Eq.~(\ref{eqn:marglikelihood}) is used for model fitting, while the second term is a complexity penalty term that maintains the Occam's razor for realizable functions as shown by~\cite{Rasmussen01occam'srazor}. The marginal likelihood involves matrix inversion and evaluating a determinant for $n\times n$ matrix, which the naive implementation would require a cubic order of computations $\mathcal{O}(n^3)$ and $\mathcal{O}(n^2)$ of storage. Approaches like Scalable Variational GP (SVGP)~\cite{hensman2013gaussian} and parametric GPR (PPGPR)~\cite{jankowiak2019parametric} have proposed approximations that lead to much better scalability.  
\section{\GP\ regression with negative datapairs}

As shown in Fig.~\ref{teaser}, we want the model to avoid certain negative datapairs in its trajectory. In other words, we want the trajectory of the Gaussian Process to have a very low probability of passing through these negative datapairs. In this section, we will first formalize the functional form of the negative datapairs and then subsequently describe our framework called \GPf\ regression.
\subsection{Definition of datapairs}
\textit{Positive datapairs}: The set of datapairs through which the \GP\ should pass are defined as positive datapairs. We assume a set of $n$ datapairs (input, positive target) with D-dimensional input vectors, $X=\{\bx_1, \cdots, \bx_n\}$ and corresponding $n\times 1$ vector of target regression values $\mathbf{y} = \{y(\mathbf{x}_1), \cdots, y(\mathbf{x}_n)\}$.

\textit{Negative datapairs}: The set of datapairs which the \GP\ should avoid (obstacles) are defined as negative datapairs. We assume a set of $m$ datapairs (input, negative target) with D-dimensional input vectors $\bar{X} = \{\bar{\bx}_1, \cdots, \bar{\bx}_m\}$ and corresponding set of negative targets $\bar{\mathbf{y}}=\{\bar{y}({\bar{\bx}_1}), \cdots, \bar{y}({\bar{\bx}_m})\}$.  
The sample value of \GP\ at input ${\bar\bx_i}$, given by $f({\bar\bx_i})$,  should be far from the negative target regression value $\bar{y}({\bar\bx_i})$. 

Note that it is possible that a particular input $\bx$ can be in both the positive and negative data pair set. This can happen, when at a particular input we want the \GP\ regression value to be close to its positive target regression value $y(\bx)$ and far from its negative target regression value $\bary(\mathbf{\bx})$.
\subsection{Functional representation of negative datapairs}
For our framework, we first get a functional representation of the negative datapairs. We define a Gaussian distribution around each of the negative datapair, $q(\bar{y}|\mathbf{\barx}) \sim \mathcal{N}(\bar{y}(\mathbf{\barx}),\sigma_\text{neg}^2)$, with mean equal to the negative target value $\bar{y}(\mathbf{x})$ and $\sigma_\text{neg}^2$ is the variance which is a hyperparameter. The Gaussian blob can also be thought of as the area of influence for the negative datapair with the variance $\sigma_\text{neg}$ indicating the spread of its influence. 

\begin{figure*}[t]
\vspace{-0.1in}
    \centering
    \includegraphics[width=\linewidth]{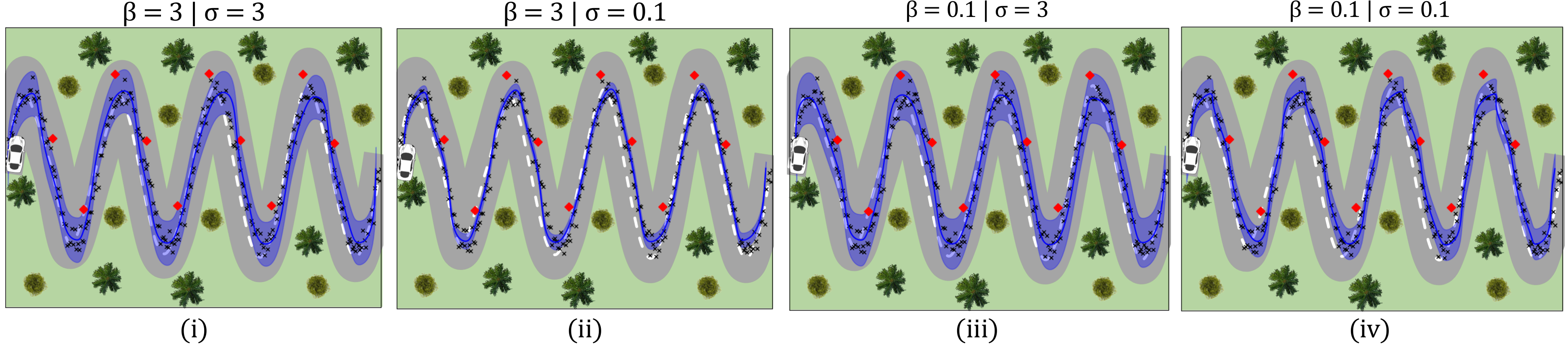}\\
    \includegraphics[width=0.6\linewidth]{figures/legend.png}\\
    \caption{Effects of hyperparameters on the trajectory prediction with \GPf\ regression framework: (i) \GPf\ framework with parameter setting as $\beta = 3$ and $\sigma_\text{neg} = 3$ (ii) \GPf\ framework with parameter setting as $\beta = 3$ and $\sigma_\text{neg} = 0.1$ (iii) \GPf\ framework with parameter setting as $\beta = 0.1$ and $\sigma_\text{neg} = 3$ (iv) \GPf\ framework with parameter setting as $\beta = 0.1$ and $\sigma_\text{neg} = 0.1$ }
    \label{fig:effects_hyp}
\end{figure*}
\subsection{\GPf\ regression framework}

The aim of our \GPf\ regression framework is to simultaneously fit the \GP\ regression on the positive datapairs ($X$) and avoid the negative datapairs ($\bar{X}$) (\ie, using them as negative constraints (NC)).
The former is achieved by maximizing the marginal likelihood given in the Eq.~(\ref{eqn:marglikelihood}). To avoid the negative datapairs, we want our \GP\ model to adjust its distribution curve so that while drawing samples from the predictive \GP\ distribution, its values do not lie in the influence region of the negative datapairs. To this end, we propose to fit the \GP\ regression model on the positive datapairs along with maximizing the Kullback-Leibler (KL) divergence between the distributions of the \GP\ regression model and the Gaussian distributions defined over the negative datapairs. 

Thus, mathematically, we want to maximize the following KL divergence given by
\begin{align}
\Delta = \text{argmax}_\theta \KL(p(\by|\theta, \bar{X})\| q(\bar{\by}|\bar{X}))
\label{eqn:kl_dist}
\end{align}
We chose to maximize the $\KL$ term in the $p\rightarrow q$ direction, as this fixes the negative datapairs distribution $q(\bar{\by}|X)$ as the reference probability distribution.
Now, since the KL divergence is an unbounded distance metric, the following section describes a practical workaround to maximize it. 
\subsection{Maximizing KL divergence using the logarithm function:}
Eq.~(\ref{eqn:kl_dist}) is increasing the distance between the \GP\ distribution and the negative datapairs distribution by maximizing the KL divergence. However, KL divergence is an unbounded function, \ie, $\KL \in [0,\infty)$. Implementing the $\KL$ divergence directly in the form of Eq.~(\ref{eqn:kl_dist}) can create problems for the gradient updates and convergence. 

In practice, to obtain a better curve fitting, we encapsulate the $\Delta$ term given by Eq.~(\ref{eqn:kl_dist}) in a logarithmic function. This ensures the \GPf\ model would focus more on fitting the positive datapairs and not solely on avoiding negative datapairs.
\vspace{-0.05in}
\subsection{\GPf: Learning and model selection}
\vspace{-0.05in}
We apply a $\log$ function to the $\KL$ given in Eq.~(\ref{eqn:kl_dist}) and write the combined objective function for our \GPf\ regression,

\begin{align}\label{eqn:likelihood-gpnc}
\begin{split}
\mathcal{L}(\theta) = \argmin_{\theta} [ &-\log p(\by|\theta, X)\\
&- \beta\log \KL(p(\by|\theta, \bar{X})\| q(\bar{\by}|\bar{X}))]
\end{split}
\end{align}
where p$(\by|\theta, X)$ is the marginal log-likelihood term that represents the model to be fitted on the observed datapoints. The parameter $\beta$ is the tradeoff hyperparameter between curve fitting and avoidance of the negative datapoints, or how relaxed can the negative constraints be.

We already know the analytical form of the log-likelihood term from Eq.~(\ref{eqn:marglikelihood}). We now focus on the $\log(\KL)$ term. Since, both the likelihood $p(\by|\theta, \bar{X})$ and negative datapair distributions $q(\bar{\by}|\bar{X})$ are modeled using Gaussians, we can simply use the analytical form of KL divergence between any two Gaussian distributions given by,
\begin{align}\label{eqn:dkl-gaussian}
\KL(p,q)= \log \frac{\sigma_2}{\sigma_1}+\frac{\sigma^2_1+(\mu_1 - \mu_2)^2}{2\sigma^2_2} -\frac{1}{2}
\end{align}
here $p,q$ are Gaussian distributions defined as $\mathcal{N}(\mu_1,\sigma_1)$ and $\mathcal{N}(\mu_2,\sigma_2)$ respectively. The $\KL$ term is adjusting the mean and variance of the likelihood $p\left(Y|\theta, \bar{X}\right)$ with respect to the fixed blobs of Gaussian distributions around the negative datapairs. Specifically, we can consider the distribution `$p\equiv\mathcal{N}(\mu_1,\sigma_1)\equiv p(\by|\theta, \bar{X})$' and `$q\equiv\mathcal{N}(\mu_2,\sigma_2)\equiv q(\bar{\by}|\bar{X})$' in Eq.~\ref{eqn:dkl-gaussian}. Now, if we refer Eq.~\ref{eqn:mean-cov}, $\mu_1 = E[\bff_*] $ and $\sigma_1 = \text{cov}(\bff_*)$ which contain the parameters $\theta$ of the \GP\ that are optimized. $\mu_2, \sigma_2$ correspond to the location of negative target $\bar{\by}$ and the hyperparameter $\sigma_\text{neg}$ of the Gaussian distribution representing the negative datapairs and are fixed. Algorithm~(\ref{alg:1}) gives an overview of the training of \GPf\ regression.

\begin{algorithm}[H]
    \SetAlgoLined
    \SetKwInOut{in}{Input}
    \SetKwInOut{param}{Parameters}
    \SetKwInOut{hparam}{Hyperparameters}
    \in{ Datapairs $\{X,\mathbf{y}\}^+$, $\{\bar{X},\mathbf{\bar{y}}\}^-$}
    \param{ \GP\ Kernel Parameters `$\theta$'}
    \hparam{ $\sigma_\text{neg}$, $\beta$}
     \While{until convergence}{
      NLL = - $p(\mathbf{y}|\theta,X)$\\
      $\theta\leftarrow$minimize (NLL)\\
      $\text{KL}_\text{div} = \beta\cdot\log \KL\left(p\left(\mathbf{\hat{y}}|\theta,\bar{X}\right)||\mathcal{N}\left(\mathbf{\bar{y}},\sigma_\text{neg}^2\right)\right)$\\
      $\theta\leftarrow$ maximize $\left(\text{KL}_\text{div}\right)$
     }
     \caption{Training of \GPf}
      \label{alg:1}
\end{algorithm}
Algorithm~(\ref{alg:1}) updates alternately  between the negative log-likelihood and KL divergence term with respect to the kernel parameters $\theta$. We empirically found that we end up with more stable solutions for $\theta$ if we update alternately instead of jointly optimizing the Eq.~\ref{eqn:likelihood-gpnc}. For different \GP\ methods we can appropriately plug-in the log-likelihood term (NLL).
\subsection{\GPf\ for scalable \& sparse \GP\ methods}
\label{sec:sparse}
In this section, we show that it is straightforward to modify the class of scalable and Sparse \GP\ regression models to account for the negative datapairs in their formulation. 

We can replace the NLL term in Algorithm~(\ref{alg:1}) by the log likelihood of the different scalable \GP\ methods. We have a scalable implementation of  the $\KL$ update, so the entire Algorithm scales well with the input data size. It is straightforward to plug-in the class of scalable and Sparse \GP\ regression models in the likelihood term of Algorithm~(\ref{alg:1}) to account for the negative datapairs in their formulation. In particular we review the SVGP model by ~\cite{hensman2013gaussian}, which is a popular scalable implementation of \GP s. We also investigate a recent parametric Gaussian Process regressors (PPGPR) method by \cite{jankowiak2019parametric}. In this section, we follow the notations given in their respective research works and give their derivations of the log likelihood function here for the sake of completeness. We evaluate the performance of these methods with our \GPf\ framework in the  experiments section.

\subsubsection{SVGP regression model}
Paper~\cite{hensman2013gaussian} proposed the Scalable Variational GP (SVGP) method. The key technical innovation was the development of inducing point methods which we now review. By introducing inducing variables $\bu$ that depend on the variational parameters $\{ \bz_m \}_{m=1}^{M}$, where $M={\rm dim}(\bu) \ll N$ and with each $\bz_m \in \mathbb{R}^d$, we augment the GP prior as $p(\bff|X) \rightarrow  p(\bff|\bu,X,Z) p(\bu|Z)$.

We then appeal to Jensen's inequality and lower bound the log joint density over the targets and inducing variables:
\begin{align}
\label{eqn:jensenenergy}
\begin{aligned}
\log p(\by, \bu |X, Z) = &\log \int d\bff p(\by|\bff) p(\bff|\bu) p(\bu)  \\
&\ge  \EE_{p(\bff|\bu)} \left[ \log p(\by|\bff)  +\log p(\bu) \right] \\
 &=  \! \sum_{i=1}^N \! \log \mathcal{N}(y_i | \bk_i^{T} K_{MM}^{-1} \bu, \sigmaobs^2) \\& - \tfrac{1}{2\sigmaobs^2}  {\rm Tr} \;\! Kt_{NN} + \log p(\bu)
 \end{aligned}
 \raisetag{69pt}
\end{align}
where $\bk_i = k(\bx_i, Z)$, $K_{MM}=k(Z,Z)$ and $Kt_{NN}$ is given by
\begin{equation}
Kt_{NN} = K_{NN} - K_{NM}  K_{MM} ^{-1} K_{MN}
\end{equation}
with $K_{NM} = K_{MN}^{\rm T} = k(X,Z)$. The essential characteristics of Eqn.~\ref{eqn:jensenenergy} are that:
i) it replaces expensive computations involving $K_{NN}$ 
with cheaper computations like $K_{MM}^{-1}$ that scale as $\OO(M^3)$; and ii) it is amenable to data subsampling,
since the log likelihood and trace terms factorize as sums over datapoints $(y_i, \bx_i)$.

\textbf{SVGP likelihood function}:
\label{sec:svgp}
SVGP proceeds by introducing a multivariate Normal variational distribution
$q(\mathbb{u}) = \mathcal{N}(\mathbb{m}, S)$.
The parameters $\mathbb{m}$ and $S$ are optimized using the ELBO (evidence lower bound), which is the expectation
of Eqn.~\ref{eqn:jensenenergy} w.r.t.~$q(\bu)$ plus an entropy term term $H[q(\bu)]$:
\begin{align}
\begin{aligned}
\label{eqn:svgp}
\mathcal{L}_{\rm svgp} & = \EE_{q(\bu)} \left[ \log p(\by, \bu |X, Z) \right] + H[q(\bu)] \\
& = \sum_{i=1}^N \left\{ \log   \mathcal{N}(y_i | \mu_\bff(\bx_i), \sigmaobs^2)
 - \frac{\sigma_\bff(\bx_i)^2}{2\sigmaobs^2} \right\} \\
 & \quad -  \KL(q(\bu) | p(\bu))
\end{aligned}
\end{align}
where KL denotes the Kullback-Leibler divergence, $\mu_\bff(\bx_i)$ is the predictive mean function given by $\mu_\bff(\bx_i) = \bk_i^{T} K_{MM}^{-1} \bmm$

and $\sigma_\bff(\bx_i)^2 \equiv \text{Var}[f_i | \bx_i] = Kt_{ii} + \bk_i^{T}  K_{MM}^{-1} S K_{MM}^{-1}  \bk_i $ denotes the latent function variance.

$\mathcal{L}_{\rm svgp}$, which depends on $\bmm, S, Z, \sigmaobs$ and the various kernel hyperparameters $\theta$,
can then be maximized with gradient methods. We refer to the resulting GP regression method as SVGP.

\subsubsection{PPGPR-ND regression model} 
Paper~\cite{jankowiak2019parametric} recently proposed a parametric Gaussian Process regressors (PPGPR) method. They assume additive Gaussian noise, $y(\mathbf{x})|f(\mathbf{x})\sim\mathcal{N}(y(\mathbf{x}); f(\mathbf{x}), \sigma^2)$, then their predictive distribution of the \GP\ evaluated at the test points $X_*$, is given by
\begin{equation}\label{eqn:mean-cov-2}
\begin{split}
    &\bff_*|X_*, X, \by, \theta, \sigma^2 \sim \mathcal{N}\left(E[\bff_*], \text{cov}(\bff_{*})\right),\\
    & E[\bff_*] = \mu_{X_{*}} + K_{X_*, X}\left[K_{X, X} + \sigma^2I\right]^{-1}\by,\\
    & \text{cov}(\bff_*) = K_{X_*, X_*} - K_{X_*, X}\left[K_{X, X}+\sigma^2I\right]^{-1}K_{X, X_*}
\end{split}
\end{equation}
We defer the reader to Section (3.2) of their paper for further details about their likelihood function.

\section{Related Works}
\begin{figure*}[t]
    \centering
    \vspace{-3mm}
    \includegraphics[width=\linewidth]
    {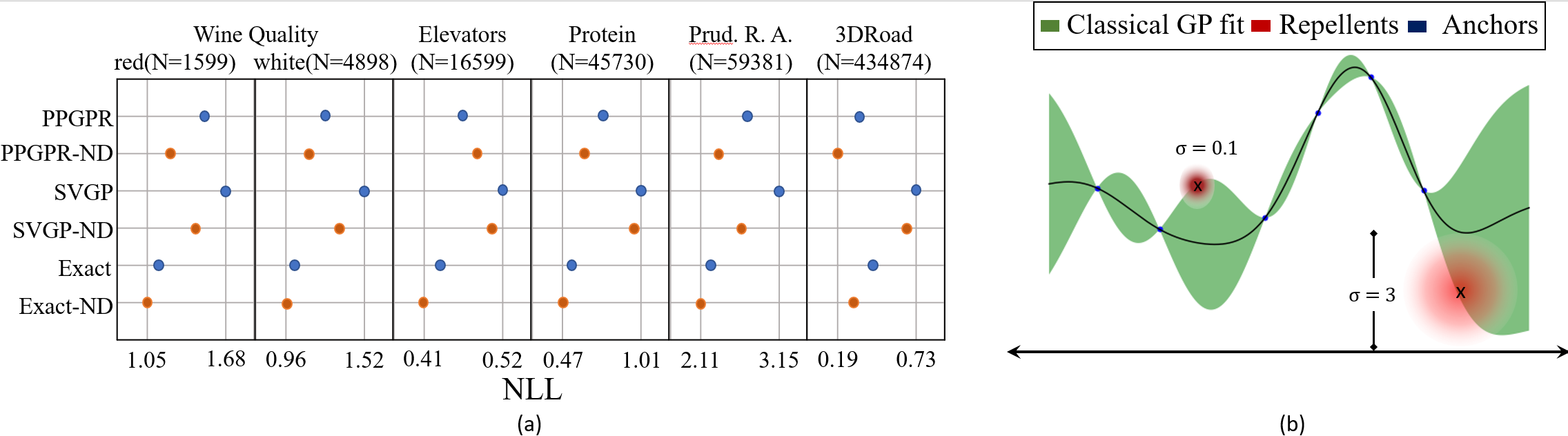}
    \vspace{-2.5mm}
    \caption{(a) Comparison on real world data: We plot test negative log-likelihoods (NLL) for 6 univariate regression datasets (lower is better). Results are averaged over 10 random train/test/valid splits. (b) Visualization of the Gaussian blobs with different $\sigma_\text{neg}$ settings. This figure conveys that though we are thinking of Gaussian blobs as a sphere of influence and intuitively expect that the larger the spread, the more it will avoid the curve, it will not always be the case. We observe that the sphere of influence is more concentrated (darker shade of red) near the marker with $\sigma_\text{neg}=0.1$ and often the \GPf \ framework will work better with lower variance values. This is due to the probability of the random sampling around the marker will also significantly go down with the increase in the value of $\sigma_\text{neg}$, thereby reducing the influence of the negative datapair on the curve. (this also explains the observations in Fig.~\ref{fig:effects_hyp}).}
    \label{fig:expt_results}
    \vspace{-0.1in}
\end{figure*}

\textit{Classical \GP}: To the best of our knowledge, the classical \GP\ regression introduced in~\cite{rasmussen2003gaussian} and many subsequent works primarily focus on positive datapairs for curve fitting. Even with the absence of the concept of negative datapairs, \GP\ regression methods have been widely used for obstacle-aware navigation task which is one of the relevant applications to evaluate our \GPf\ framework.

\textit{\GP s for navigation}: \GP s are extensively used in the field of navigation and often are a component of path planning algorithms. \cite{ellis2009modelling} used \GP\ regression in modeling the pedestrian trajectories by using positive datapairs. \cite{aoude2013probabilistically} used a heuristic-based approach over \GP\ regression to incorporate dynamic changes and environmental constraint in the surroundings. Their solution, named RR-GP, builds a learned motion pattern model by combining the flexibility of \GP\ with the efficiency of RRTReach, a sampling-based reachability computation. Obstacle trajectory GP predictions are conditioned on dynamically feasible paths identified from the reachability analysis, yielding more accurate predictions of future behavior. \cite{8500614} introduced the use of \GP\ regression for long-term location prediction for collision avoidance in Connected Vehicle (CV) environment. The \GP s are used to model the trajectory of the vehicles using the historical data. The collected data from vehicles together with GPR models received from infrastructure are then used to predict the future trajectories of vehicles in the scene. \cite{meera2019obstacleaware} designed an Obstacle-aware Adaptive Informative Path Planning (OA-IPP) algorithm for target search in
cluttered environments using UAVs. This method uses \GP\ to detect the obstacles/target, which the UAV gets by marking a dense number of points (positive datapairs) around the obstacles. \cite{hewing2020simulation,yuan2019diverse} are some of the works using sampling-based techniques for trajectory prediction.
\cite{7139222} is one of the work which tries to incorporate the concept of negative datapairs in classical \GP\ construct by introducing a \textit{leveraged parameter} in kernel function. The authors demonstrate that having the ability to incorporate negative targets increases the efficiency of trajectory predictions. However, this approach fundamentally differs from ours in terms of incorporation of negative datapairs; ours try to maximize the KL-Divergence between $\hat{y}(x)$ and $\bar{y}(x)$ while theirs utilizes additional \textit{leveraged parameter} in the kernel function. Besides, in our approach, the covariance matrix relies only on the positive datapair in contrast to the approach put forth by \cite{7139222} where negative datapairs are also incorporated in the covariance matrix to optimize the kernel parameters. Our approach is more scalable as the covariance matrix's size does not increase to incorporate the negative datapairs.

\textit{Scalable \GP}: Na\"ive implementation of \GP\ regression is not scalable for large datasets as the model selection and inference of \GP\ requires a cubic order of computations $\mathcal{O}(n^3)$ and $\mathcal{O}(n^2)$ of storage. Since the \GPf\ framework is quite generic and can work for various scale \GP\ methods, we highlight a few of them here. \cite{hensman2013gaussian,dai2014gaussian,gal2014distributed} are some of the known scalable methods suitable for our framework as they use stochastic gradient descent for optimization. Furthermore, recent works~\cite{wilson2015kernel,wilson2015thoughts,wilson2016stochastic} have improved scalability by reducing the learning to $\mathcal{O}(n)$ and test prediction to $\mathcal{O}(1)$ under some assumptions.

\textit{Negative datapairs in other domains}: The concept of negative datapairs has been extensively utilized in the self-supervised learning. Applications include learning word embeddings~\cite{NIPS2013_5021,mnih2013learning}, image representations~\cite{he2020momentum,misra2020self,Feng_2019_CVPR}, video representations~\cite{8462891,Fernando_2017_CVPR,misra2016shuffle,harley2020learning}, etc. In these works, negative and positive samples are created as pseudo labels to train a neural network to learn the deep representations of the inputs.

\textit{Jointly optimizing multiple loss functions.} \GPf\ jointly optimize 2 loss functions, refer Eq.~\ref{eqn:likelihood-gpnc}. We acknowledge that there can be multiple ways to balance between multiple loss functions. The process followed by~\cite{rajbhandari2019antman} to balance between different losses and finding optimal hyperparameter values have shown to produce improved results. An alternative way to include negative datapoints can be to model the confusion matrix (true positives, false negatives, etc.) itself. This approach is mainly used for classification tasks, refer~\cite{shrivastava2020grnular,shrivastava2022grnular}, where the deep learning model was optimized for F-beta score.

\section{Experiments}\label{sec:expts}
We compared various \GP\ regression models in their classical form (using only positive datapairs) with their corresponding \GPf\ regression models where we used our negative constraints framework. We used Negative Log-likelihood (NLL) and Root Mean Squared Error (RMSE) as our evaluation metrics. We compared our framework on a synthetic dataset and six real world datasets. Throughout our experiments, we found that for every \GP\ model, the \GPf\ regression framework outperforms its corresponding classical \GP\ regression setting. We used GPytorch~\cite{gardner2018gpytorch} to implement all the \GP\ (ours + baselines) models. We use zero mean value and the RBF kernel for \GP\ prior for all of the models unless mentioned otherwise.

\subsection{Trajectory prediction using \GPf\ }
\begin{table*}[t]
\vspace{-2mm}
\renewcommand{\tabcolsep}{6pt}
\renewcommand{\arraystretch}{1}
\footnotesize
  \centering
  \caption{\textbf{Runtime comparison} of the classical \GP\ and \GPf\ frameworks which includes negative datapairs on different datasets. $\Delta_t$ is the runtime difference of the \GP\ model in \GPf\ framework vs the classical \GP\ framework. We used GPU accelerated \GP\ implementation of GPyTorch library.}
  \vspace{-.1in}
  \begin{tabular}{@{}lccccc@{}}
    \toprule
    Datasets & Size of data&Type of Target Variable &  $\Delta_t$ Exact \GP\  & $\Delta_t$ sparse SVGP \\
    \midrule
    Wine quality - red &1599& Discrete  &   21ms   &  3.1s \\
    Wine quality - white &4898&Discrete    & 40ms  &  3.7s\\
   Elevators& 16599& Continuous  & 5s & 34s\\
    Protein&45730&Continuous& 50s & 37s\\
     Prudential&59381&Discrete& 1.2 s & 34s\\
     3DRoad&434874&Continuous& 2208s & 292s\\
    \bottomrule
  \end{tabular}
  \label{tab:results}
\end{table*}
This set of experiments are inspired by~\cite{ellis2009modelling}, we synthesize a 2d-virtual traffic scene given by Fig.~\ref{fig:vir_scene}. Furthermore, the road contain pitfalls, roadblocks, accidents, etc., that need to be avoided are represented as red diamonds in the figure. We designate these targets as negative targets that are to be avoided for ensuring the safety of traffic. We want to model an agent's trajectory using a \GP\ regression model such that it takes the agent's present location $(x,y)$ as input and predicts the probability of the agent's next location $(\hat{x},\hat{y})$. There are a total of 10 negative datapairs present in the scene. We then sample 250 observed co-ordinates on the 2d virtual path for modeling the future trajectory of the navigating agent.
 
To evaluate our method, we trained two baselines, namely a classical \GP\ model and an NS\GP\  model~\cite{7139222} in addition to ours \GPf\ model for predicting the trajectory of the agent. For all the models, we utilize a constant mean and an RBF kernel for the \GP\ prior. All the models were trained for 100 epochs. It can be observed from Fig.~\ref{fig:vir_scene}.a that the classical \GP\ framework lacks the ability to incorporate negative datapairs, which results in a loosely fitted \GP\ model. We also evaluate against an existing approach NS\GP\ that incorporates the additional constraints in the form of negative datapairs. It can be observed in Fig.~\ref{fig:vir_scene}.b. that this approach does repel the curve from negative datapairs. However, the fit of the NS\GP\ model as observed by the figure is erratic in nature, and at times the predicted trajectory goes out of the environmental constraints, which can put the car in danger. 

All these baselines, when juxtaposed with our model, demonstrate that our model \GPf\ outperforms both of them in terms of a better curve-fit and avoidance of all the negative datapairs as shown in Fig~\ref{fig:vir_scene}.c.  We assert that for a critical safety application such as navigation, \GPf\ is superior to the NS\GP\ and classical \GP\ in incorporating the negative constraints.

\textbf{Understanding the hyperparameters:}\label{sec:hyp}
We performed a few additional experiments on the same simulated scene with different set of hyperparameters to understand their effects. These experiments with different settings of hyperparameter $\beta$ and $\sigma_\text{neg}$ can be observed in Fig.~\ref{fig:effects_hyp}. Moreover, in these set of experiments it is easy to demonstrate the impact of the hyperparameters value in Eq.~(\ref{eqn:likelihood-gpnc}) on the \GP\ regression. As observed from the Fig~\ref{fig:effects_hyp}.i vs. Fig~\ref{fig:effects_hyp}.iii and Fig~\ref{fig:effects_hyp}.ii vs. Fig~\ref{fig:effects_hyp}.iv decreasing the value of $\beta$ results in reduction of influence of the negative datapairs. Another observation that can be made is the influence of negative points reduces if there is an increase in the values of $\sigma_\text{neg}$ as seen in Fig~\ref{fig:effects_hyp}.i vs. Fig~\ref{fig:effects_hyp}.ii and Fig~\ref{fig:effects_hyp}.iii vs. Fig~\ref{fig:effects_hyp}.iv. 

The experiment discussed in $\S$\ref{sec:hyp} demonstrates that lowering the $\beta$ or increasing the $\sigma_\text{neg}$ generate a similar effect, \ie, it lowers the influence of negative datapairs in curve fitting of \GPf .  The effect of parameter $\beta$ is fairly straightforward to understand as decreasing the value of $\beta$ would result in the decline of the influence of the $\KL$ term, which constitutes a negative constraint on the Gaussian likelihood given by the Eq.~(\ref{eqn:likelihood-gpnc}). Fig.\ref{fig:effects_hyp}.ii and Fig.\ref{fig:effects_hyp}.iv demonstrate the influence of parameter $\beta$ well on curve-fitting of \GPf\ .

The effect of $\sigma_\text{neg}$ on the curve-fitting of \GPf\ is however interesting and needs to be carefully understood. We observe the loss function given in Eq.~(\ref{eqn:dkl-gaussian}) and more specifically the analytical form of KL-divergence for Gaussian distribution given by Eq.~(\ref{eqn:dkl-gaussian}). If we consider the terms other than $\sigma_2=\sigma_\text{neg}$ in Eq.~(\ref{eqn:dkl-gaussian}) as constant, we get a function of the form $\log(x)+\frac{1}{x^2}$. We thus need to carefully choose the $\sigma_\text{neg}$ value to be before the knee point of the curve, so as to control the $\text{KL}_\text{div}$ term for better curve fitting.

Fig.~(\ref{fig:expt_results}.b) illustrates that if we keep the $\sigma_\text{neg}$ values low, the spread of the Gaussian blob is less. Nevertheless, the probability of random sampling over the region (the small Gaussian blob) would be extremely concentrated at the marker because of the lower value of $\sigma_\text{neg}$. This would result in higher probability values of the random samples over the region closer to the vicinity of a negative target for smaller Gaussian blobs as compared to the big ones.

\begin{figure*}[t]
    \centering
    \includegraphics[width=0.99\linewidth]{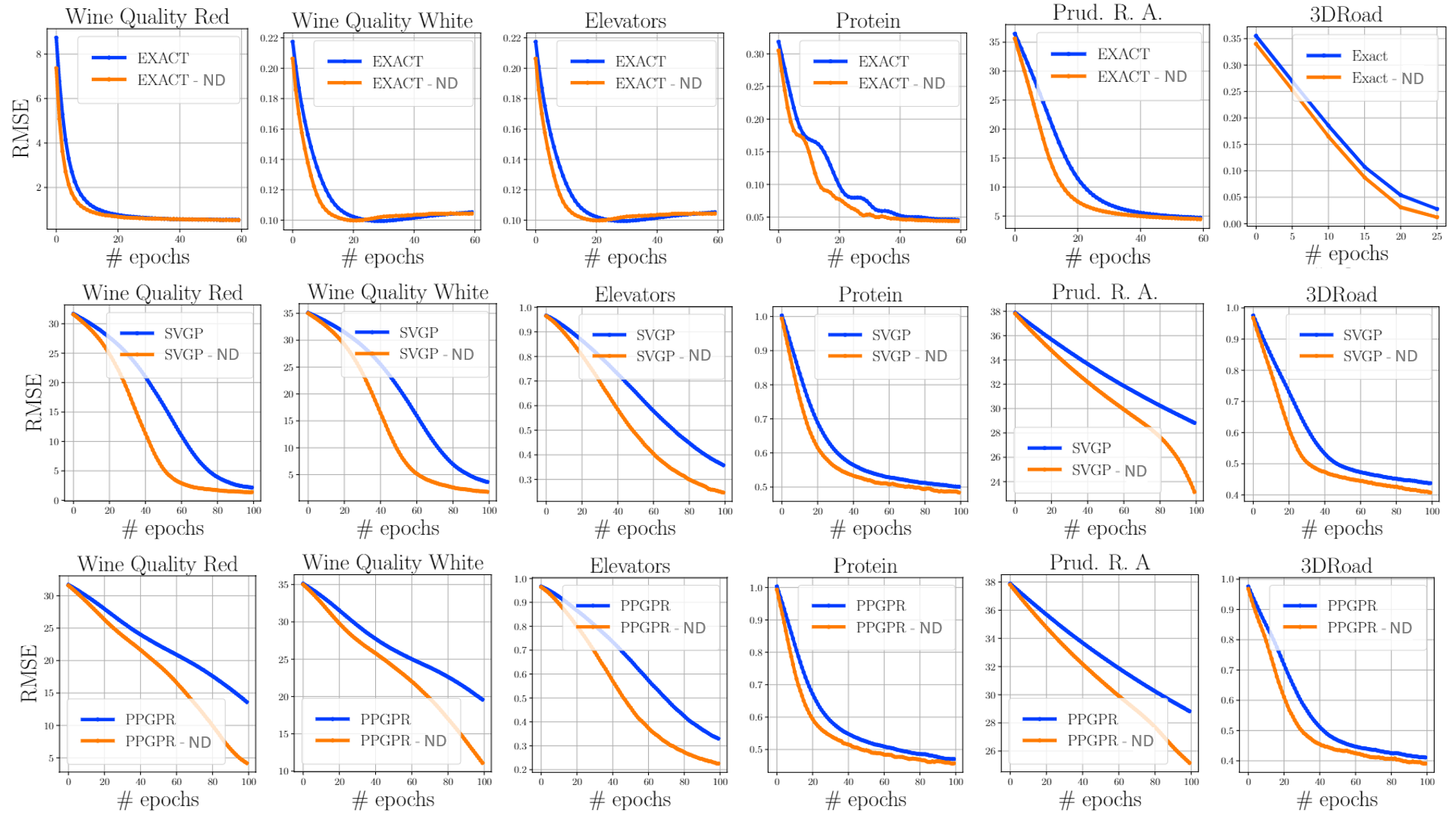}
    \caption{RMSE plots on real world data (top - \textbf{Exact \GP}; middle - \textbf{SVGP}; bottom - \textbf{PPGPR}): Plots show the test RMSE for six univariate regression datasets (lower is better). Models are fitted by using cross validation on training data. Convergence of \GPf\ framework is consistently faster than its classical \GP\ counterpart for all the models.
    }
    \label{fig:rmse-plots}
    \vspace{-0.1in}
\end{figure*}

\subsection{Realworld datasets}
We evaluated our \GPf\ framework on six real world datasets with the number of datapoints ranging from $N\sim 1500,5000,15000, 50000,450000$ and the number of input dimension $d\in[3,127]$. Among the six datasets five of them are from UCI repository~\cite{Dua:2019} (Wine quality - red, white, Elevators, Protein, and 3DRoad), while the sixth one is from Kaggle Competition (Prudential life insurance risk assessment). These datasets consists of two different kind of prediction/regression variables namely discrete variable and continuous variable. For discreet variable the value of elements lies between certain range, \ie, integer values lie between $[0,10]$. For continuous variable, the value of target regression can be any real number. Datasets (Prudential risk assessment, Wine quality - red, white) are all discrete target variable datasets while datasets (Elevators, Protein and 3DRoad) have continuous target variables.

\textbf{Random shuffling technique: }
We use Random shuffling technique for creating negative datapairs for our \GPf\ framework.As we are only given positive regression target values in these datasets, we create pseudo-negative regression targets by randomly shuffling the labels and pairing them with the input points $x$ to create negative datapairs. This process leads to the creation of additional data, which can be augmented with a set of valid datapairs. 
Mathematically, a valid datapair is defined as any input $\mathbf{x}$ correctly associated with true target value/label $y(\bx)_\text{True}$ and a valid negative datapair is defined as any input $\mathbf{\bar{x}}$ associated with randomly shuffled target (a wrong target!) $y(\mathbf{x})_\text{Shuffled}$. It is important to note that \vspace{-.05in}$$y(\bx)_\text{True} \ne y(\mathbf{x})_\text{Shuffled}$$\vspace{-.3in}

We compared our model against the standard baselines of Exact \GP ~\cite{gardner2019gpytorch,wang2019exact}, and Sparse \GP\ methods like SVGP~\cite{hensman2013gaussian} and PPGPR~\cite{jankowiak2019parametric}. For training the sparse methods, we used 1000 inducing points. We trained the models using Adam optimizer with a learning rate of 0.1 for 400 epochs on each dataset. For \GPf\ framework, we used 200 negative datapairs obtained from random shuffling technique. We maintained consistency across all the models in terms of maintaining a constant mean and RBF kernel for a \GP\ prior. 

Fig.~\ref{fig:expt_results} compares negative log likelihood values of various \GP\ regression methods, in both classical and \GPf\ frameworks, on all six real datasets. The orange dots represents our methods while the blue dots in the plots depict the baselines. It can be observed that \GPf\ framework outperforms the classical \GP\ framework. Methods like SVGP - ND and Exact \GP\ - ND performs on an average 0.2 nats better than baseline SVGP and Exact \GP\ respectively. It is interesting to note that including negative datapairs by the `random shuffling technique' is quite effective and we can observe gains in terms of model performance.

Figure \ref{fig:rmse-plots} shows the RMSE plots for Exact \GP , SVGP, and PPGPR models with the classical \GP\ juxtaposed on to the models with the negative constraint \GPf\ framework. 
The plots show that our \GPf\ models converge faster and better than the classical \GP\ models for the six univariate real-world datasets. Also, with the increase in data size, the convergence curve of the \GPf\ model becomes steeper.

\textbf{Runtime of \GPf\ Framework:} Table~\ref{tab:results} compares runtime difference between the classical \GP\ and the \GPf\ to see how the KL divergence term that models negative datapairs affects the runtime of our framework. We train all the \GP\ models for 50 epochs and measure the average excess time ($\Delta_t$) value over 10 runs. It can be observed from Table~\ref{tab:results} that the $\Delta_t$ depends on the size of the dataset and also on the type of target regression variable. For the Exact \GP\ and discrete target variables setting the $\Delta_t$ does not increase much with size of the dataset, however for continuous target variable there is a considerable amount of increase. For SVGP model the increase in $\Delta_t$ can be attributed to the size of the dataset. Overall, the average increase in training is not very significant. Thus, our experiments indicate that the added penalty term to likelihood term in Eq.~(\ref{eqn:likelihood-gpnc}) does not significantly affect the \textbf{scalability} of current scalable \GP\ architectures.

\textit{Handling imbalanced data}: Often in real datasets, we encounter imbalance in the data which means the ratio of positive to negative samples is quite skewed. Popular techniques like SMOTE~\cite{chawla2002smote} that augments synthetic data to help with training can be extended to our framework. Other techniques that adjust the learning objective function to account for class imbalance handling like~\cite{rahman2013addressing,bhattacharya2017icu,bhattacharya2019methods,shrivastava2015classification,shrivastava2021system} suggests good modifications that can be easily incorporated in the \GPf\ loss.

\section{Interesting Applications for \GPf\ }
Apart from the trajectory prediction and standard time-series modeling tasks that traditionally leverage Gaussian Processes, we list out some other potentially interesting applications for our \GPf\ framework. We hope that this will encourage readers to explore and discover more innovative applications.


\textit{Study time-varying Graph trajectories.} We have started early exploration on combining \GPf\ with methods that recover Conditional Independence (CI) graphs to understand dynamics of a system~\cite{hallac2015network,shrivastava2022methods}. CI graphs are based on solving the graphical lasso problem or its variants that models the input data as a multivariate Gaussian distribution. since, the \GP's and CI graph recovery methods~\cite{friedman2008sparse,rolfs2012iterative}, including recently developed deep learning methods~\cite{shrivastava2019glad,shrivastava2022uglad,shrivastava2022a} share the same underlying Gaussian distribution assumptions, one can look into functional traits for defining a temporal process. So, at every timestep, we will have a CI graph that shows how the features are connected and the fitting a \GPf\ over these time-varying graphs can help in predicting the future variations by including system constraints. Such problems can be found in modeling trajectory-based differential expression analysis for single-cell RNA sequencing data and ensemble gene regulatory network prediction tasks~\cite{lin2019continuous,fischer2019inferring,aluru2021engrain}. 

\textit{Video generation} recent works on video generation, given a few context frames~\cite{shrivastava2021diverse,shrivastava2021diversethesis,shrivastava2024thesis,bodla2021hierarchical,shrivastava2025video,shrivastava2023video,shrivastava2024video} can greatly benefit by incorporating negative datapairs modeling by our \GPf\ framework. 

\textit{Merging with Probabilistic Graphical models.} PGMs are powerful tools to represent functional dependencies of a systems attributes or features~\cite{koller2009probabilistic}. Although, there has been considerable progress in enriching the functional representation of the PGMs, Bayesian Networks~\cite{welling2011bayesian,zheng2018dags,zheng2020learning} \& Markov networks~\cite{belilovsky2017learning,pu2021learning,shrivastava2022neural}, much research is needed to connect these PGMs obtained at each time-step. We believe that \GPf\ can provide a necessary inductive bias~\cite{shrivastava2019cooperative,shrivastava2020using} to connect the PGMs over the time-series. Such systems will find applications in understanding climate forecasting, speech recognition, audio understanding etc.~\cite{sun2014monthly,saini2022recognizing,roche2017valorcarn,fize2017geodict,wilkinson2019gaussian} to list a few.

\section{Conclusion}
We presented a novel and generic Gaussian Process regression framework \GPf\ that incorporates negative constraints along with fitting the curve for the positive datapairs. Our key idea was to assume small blobs of Gaussian distribution on the negative datapairs. Then, while fitting the \GP\ regression on the positive datapairs, our \GPf\ framework simultaneously maximizes the KL divergence from the negative datapairs. Our work highlights the benefits of modeling the negative datapairs for \GP s and our experiments support the effectiveness of our approach. We hope that this successful realization of the concept of negative datapairs for \GP\ regression will be useful in variety of applications.

\bibliography{all_citations}

\begin{thebibliography}{10}

\bibitem{aluru2021engrain}
Maneesha Aluru, Harsh Shrivastava, Sriram~P Chockalingam, Shruti Shivakumar, and Srinivas Aluru.
\newblock {EnGRaiN}: a supervised ensemble learning method for recovery of large-scale gene regulatory networks.
\newblock {\em Bioinformatics}, 2021.

\bibitem{aoude2013probabilistically}
Georges~S Aoude, Brandon~D Luders, Joshua~M Joseph, Nicholas Roy, and Jonathan~P How.
\newblock Probabilistically safe motion planning to avoid dynamic obstacles with uncertain motion patterns.
\newblock {\em Autonomous Robots}, 35(1):51--76, 2013.

\bibitem{belilovsky2017learning}
Eugene Belilovsky, Kyle Kastner, Ga{\"e}l Varoquaux, and Matthew~B Blaschko.
\newblock Learning to discover sparse graphical models.
\newblock In {\em International Conference on Machine Learning}, pages 440--448. PMLR, 2017.

\bibitem{bhattacharya2017icu}
Sakyajit Bhattacharya, Vaibhav Rajan, and Harsh Shrivastava.
\newblock {ICU} mortality prediction: a classification algorithm for imbalanced datasets.
\newblock In {\em Proceedings of the AAAI Conference on Artificial Intelligence}, volume~31, 2017.

\bibitem{bhattacharya2019methods}
Sakyajit Bhattacharya, Vaibhav Rajan, and Harsh Shrivastava.
\newblock Methods and systems for predicting mortality of a patient, November~5 2019.
\newblock US Patent 10,463,312.

\bibitem{bodla2021hierarchical}
Navaneeth Bodla, Gaurav Shrivastava, Rama Chellappa, and Abhinav Shrivastava.
\newblock Hierarchical video prediction using relational layouts for human-object interactions.
\newblock In {\em Proceedings of the IEEE/CVF Conference on Computer Vision and Pattern Recognition}, pages 12146--12155, 2021.

\bibitem{chawla2002smote}
Nitesh~V Chawla, Kevin~W Bowyer, Lawrence~O Hall, and W~Philip Kegelmeyer.
\newblock {SMOTE}: synthetic minority over-sampling technique.
\newblock {\em Journal of artificial intelligence research}, 16:321--357, 2002.

\bibitem{7139222}
S.~{Choi}, E.~{Kim}, K.~{Lee}, and S.~{Oh}.
\newblock Leveraged non-stationary gaussian process regression for autonomous robot navigation.
\newblock In {\em 2015 IEEE International Conference on Robotics and Automation (ICRA)}, pages 473--478, 2015.

\bibitem{dai2014gaussian}
Zhenwen Dai, Andreas Damianou, James Hensman, and Neil Lawrence.
\newblock Gaussian process models with parallelization and gpu acceleration.
\newblock In {\em arXiv}, 2014.

\bibitem{Dua:2019}
Dheeru Dua and Casey Graff.
\newblock {UCI} machine learning repository, 2017.

\bibitem{ellis2009modelling}
David Ellis, Eric Sommerlade, and Ian Reid.
\newblock Modelling pedestrian trajectory patterns with gaussian processes.
\newblock In {\em 2009 IEEE 12th International Conference on Computer Vision Workshops, ICCV Workshops}, pages 1229--1234. IEEE, 2009.

\bibitem{Feng_2019_CVPR}
Zeyu Feng, Chang Xu, and Dacheng Tao.
\newblock Self-supervised representation learning by rotation feature decoupling.
\newblock In {\em Proceedings of the IEEE/CVF Conference on Computer Vision and Pattern Recognition (CVPR)}, June 2019.

\bibitem{Fernando_2017_CVPR}
Basura Fernando, Hakan Bilen, Efstratios Gavves, and Stephen Gould.
\newblock Self-supervised video representation learning with odd-one-out networks.
\newblock In {\em Proceedings of the IEEE Conference on Computer Vision and Pattern Recognition (CVPR)}, July 2017.

\bibitem{fischer2019inferring}
David~S Fischer, Anna~K Fiedler, Eric~M Kernfeld, Ryan~MJ Genga, Aim{\'e}e Bastidas-Ponce, Mostafa Bakhti, Heiko Lickert, Jan Hasenauer, Rene Maehr, and Fabian~J Theis.
\newblock Inferring population dynamics from single-cell rna-sequencing time series data.
\newblock {\em Nature biotechnology}, 37(4):461--468, 2019.

\bibitem{fize2017geodict}
Jacques Fize, Gaurav Shrivastava, and Pierre~Andr{\'e} M{\'e}nard.
\newblock Geodict: an integrated gazetteer.
\newblock In {\em Proceedings of Language, Ontology, Terminology and Knowledge Structures Workshop (LOTKS 2017)}, 2017.

\bibitem{friedman2008sparse}
Jerome Friedman, Trevor Hastie, and Robert Tibshirani.
\newblock Sparse inverse covariance estimation with the graphical lasso.
\newblock {\em Biostatistics}, 9(3):432--441, 2008.

\bibitem{gal2014distributed}
Yarin Gal, Mark van~der Wilk, and Carl~E. Rasmussen.
\newblock Distributed variational inference in sparse gaussian process regression and latent variable models.
\newblock In {\em arXiv}, 2014.

\bibitem{gardner2018gpytorch}
Jacob~R Gardner, Geoff Pleiss, David Bindel, Kilian~Q Weinberger, and Andrew~Gordon Wilson.
\newblock Gpytorch: Blackbox matrix-matrix gaussian process inference with gpu acceleration.
\newblock In {\em Advances in Neural Information Processing Systems}, 2018.

\bibitem{gardner2019gpytorch}
Jacob~R. Gardner, Geoff Pleiss, David Bindel, Kilian~Q. Weinberger, and Andrew~Gordon Wilson.
\newblock Gpytorch: Blackbox matrix-matrix gaussian process inference with gpu acceleration.
\newblock In {\em arXiv}, 2019.

\bibitem{8500614}
S.~A. {Goli}, B.~H. {Far}, and A.~O. {Fapojuwo}.
\newblock Vehicle trajectory prediction with gaussian process regression in connected vehicle environment$\star$.
\newblock In {\em 2018 IEEE Intelligent Vehicles Symposium (IV)}, pages 550--555, 2018.

\bibitem{hallac2015network}
David Hallac, Jure Leskovec, and Stephen Boyd.
\newblock Network lasso: Clustering and optimization in large graphs.
\newblock In {\em Proceedings of the 21th ACM SIGKDD international conference on knowledge discovery and data mining}, pages 387--396, 2015.

\bibitem{handcock1993bayesian}
Mark~S Handcock and Michael~L Stein.
\newblock A bayesian analysis of kriging.
\newblock {\em Technometrics}, 35(4):403--410, 1993.

\bibitem{harley2020learning}
Adam~W. Harley, Shrinidhi~K. Lakshmikanth, Fangyu Li, Xian Zhou, Hsiao-Yu~Fish Tung, and Katerina Fragkiadaki.
\newblock Learning from unlabelled videos using contrastive predictive neural 3d mapping.
\newblock In {\em arXiv}, 2020.

\bibitem{he2020momentum}
Kaiming He, Haoqi Fan, Yuxin Wu, Saining Xie, and Ross Girshick.
\newblock Momentum contrast for unsupervised visual representation learning.
\newblock In {\em arXiv}, 2020.

\bibitem{hensman2013gaussian}
James Hensman, Nicolo Fusi, and Neil~D. Lawrence.
\newblock Gaussian processes for big data.
\newblock In {\em arXiv}, 2013.

\bibitem{hewing2020simulation}
Lukas Hewing, Elena Arcari, Lukas~P Fr{\"o}hlich, and Melanie~N Zeilinger.
\newblock On simulation and trajectory prediction with gaussian process dynamics.
\newblock In {\em Learning for Dynamics and Control}, pages 424--434. PMLR, 2020.

\bibitem{jankowiak2019parametric}
Martin Jankowiak, Geoff Pleiss, and Jacob~R Gardner.
\newblock Parametric gaussian process regressors.
\newblock {\em arXiv}, pages arXiv--1910, 2019.

\bibitem{jazwinski2007stochastic}
Andrew~H Jazwinski.
\newblock {\em Stochastic processes and filtering theory}.
\newblock Courier Corporation, 2007.

\bibitem{koller2009probabilistic}
Daphne Koller and Nir Friedman.
\newblock {\em Probabilistic graphical models: principles and techniques}.
\newblock MIT press, 2009.

\bibitem{lin2019continuous}
Chieh Lin and Ziv Bar-Joseph.
\newblock Continuous-state hmms for modeling time-series single-cell rna-seq data.
\newblock {\em Bioinformatics}, 35(22):4707--4715, 2019.

\bibitem{meera2019obstacleaware}
Ajith~Anil Meera, Marija Popovic, Alexander Millane, and Roland Siegwart.
\newblock Obstacle-aware adaptive informative path planning for uav-based target search.
\newblock In {\em arXiv.cs.RO}, 2019.

\bibitem{NIPS2013_5021}
Tomas Mikolov, Ilya Sutskever, Kai Chen, Greg~S Corrado, and Jeff Dean.
\newblock Distributed representations of words and phrases and their compositionality.
\newblock In C.~J.~C. Burges, L.~Bottou, M.~Welling, Z.~Ghahramani, and K.~Q. Weinberger, editors, {\em Advances in Neural Information Processing Systems 26}, pages 3111--3119. Curran Associates, Inc., 2013.

\bibitem{misra2020self}
Ishan Misra and Laurens van~der Maaten.
\newblock Self-supervised learning of pretext-invariant representations.
\newblock In {\em Proceedings of the IEEE/CVF Conference on Computer Vision and Pattern Recognition}, pages 6707--6717, 2020.

\bibitem{misra2016shuffle}
Ishan Misra, C~Lawrence Zitnick, and Martial Hebert.
\newblock Shuffle and learn: unsupervised learning using temporal order verification.
\newblock In {\em European Conference on Computer Vision}, pages 527--544. Springer, 2016.

\bibitem{mnih2013learning}
Andriy Mnih and Koray Kavukcuoglu.
\newblock Learning word embeddings efficiently with noise-contrastive estimation.
\newblock In {\em Advances in neural information processing systems}, pages 2265--2273, 2013.

\bibitem{pu2021learning}
Xingyue Pu, Tianyue Cao, Xiaoyun Zhang, Xiaowen Dong, and Siheng Chen.
\newblock Learning to learn graph topologies.
\newblock {\em Advances in Neural Information Processing Systems}, 34, 2021.

\bibitem{rahman2013addressing}
M~Mostafizur Rahman and Darryl~N Davis.
\newblock Addressing the class imbalance problem in medical datasets.
\newblock {\em International Journal of Machine Learning and Computing}, 3(2):224, 2013.

\bibitem{rajbhandari2019antman}
Samyam Rajbhandari, Harsh Shrivastava, and Yuxiong He.
\newblock Antman: Sparse low-rank compression to accelerate rnn inference.
\newblock {\em arXiv preprint arXiv:1910.01740}, 2019.

\bibitem{rasmussen2003gaussian}
Carl~Edward Rasmussen.
\newblock Gaussian processes in machine learning.
\newblock In {\em Summer school on machine learning}, pages 63--71. Springer, 2003.

\bibitem{Rasmussen01occam'srazor}
Carl~Edward Rasmussen and Zoubin Ghahramani.
\newblock Occam's razor.
\newblock In {\em In Advances in Neural Information Processing Systems 13}, pages 294--300. MIT Press, 2001.

\bibitem{roche2017valorcarn}
Mathieu Roche, Maguelonne Teisseire, and Gaurav Shrivastava.
\newblock Valorcarn-tetis: Terms extracted with biotex.
\newblock 2017.

\bibitem{rolfs2012iterative}
Benjamin Rolfs, Bala Rajaratnam, Dominique Guillot, Ian Wong, and Arian Maleki.
\newblock Iterative thresholding algorithm for sparse inverse covariance estimation.
\newblock {\em Advances in Neural Information Processing Systems}, 25:1574--1582, 2012.

\bibitem{saini2022recognizing}
Nirat Saini, Bo~He, Gaurav Shrivastava, Sai~Saketh Rambhatla, and Abhinav Shrivastava.
\newblock Recognizing actions using object states.
\newblock In {\em ICLR2022 Workshop on the Elements of Reasoning: Objects, Structure and Causality}, 2022.

\bibitem{8462891}
P.~{Sermanet}, C.~{Lynch}, Y.~{Chebotar}, J.~{Hsu}, E.~{Jang}, S.~{Schaal}, S.~{Levine}, and G.~{Brain}.
\newblock Time-contrastive networks: Self-supervised learning from video.
\newblock In {\em 2018 IEEE International Conference on Robotics and Automation (ICRA)}, pages 1134--1141, 2018.

\bibitem{shrivastava2021diversethesis}
Gaurav Shrivastava.
\newblock {\em Diverse Video Generation}.
\newblock PhD thesis, University of Maryland, College Park, 2021.

\bibitem{shrivastava2024thesis}
Gaurav Shrivastava.
\newblock {\em Advanced video modeling techniques for generation and enhancement tasks}.
\newblock PhD thesis, University of Maryland, College Park, 2024.

\bibitem{shrivastava2023video}
Gaurav Shrivastava, Ser-Nam Lim, and Abhinav Shrivastava.
\newblock Video dynamics prior: An internal learning approach for robust video enhancements.
\newblock In {\em Thirty-seventh Conference on Neural Information Processing Systems}, 2023.

\bibitem{shrivastava2024video}
Gaurav Shrivastava, Ser-Nam Lim, and Abhinav Shrivastava.
\newblock Video decomposition prior: Editing videos layer by layer.
\newblock In {\em The Twelfth International Conference on Learning Representations}, 2024.

\bibitem{shrivastava2021diverse}
Gaurav Shrivastava and Abhinav Shrivastava.
\newblock Diverse video generation using a {Gaussian} process trigger.
\newblock {\em arXiv preprint arXiv:2107.04619}, 2021.

\bibitem{shrivastava2025video}
Gaurav Shrivastava and Abhinav Shrivastava.
\newblock Video prediction by modeling videos as continuous multi-dimensional processes.
\newblock In {\em Proceedings of the IEEE/CVF Conference on Computer Vision and Pattern Recognition}, pages 7236--7245, 2024.

\bibitem{shrivastava2020using}
Harsh Shrivastava.
\newblock {\em On Using Inductive Biases for Designing Deep Learning Architectures}.
\newblock PhD thesis, Georgia Institute of Technology, 2020.

\bibitem{shrivastava2019cooperative}
Harsh Shrivastava, Eugene Bart, Bob Price, Hanjun Dai, Bo~Dai, and Srinivas Aluru.
\newblock Cooperative neural networks ({CoNN}): Exploiting prior independence structure for improved classification.
\newblock {\em arXiv preprint arXiv:1906.00291}, 2019.

\bibitem{shrivastava2022methods}
Harsh Shrivastava and Urszula Chajewska.
\newblock Methods for recovering conditional independence graphs: A survey.
\newblock {\em arXiv preprint arXiv:2211.06829}, 2022.

\bibitem{shrivastava2022neural}
Harsh Shrivastava and Urszula Chajewska.
\newblock Neural graphical models.
\newblock {\em arXiv preprint arXiv:2210.00453}, 2022.

\bibitem{shrivastava2022a}
Harsh Shrivastava, Urszula Chajewska, Robin Abraham, and Xinshi Chen.
\newblock A deep learning approach to recover conditional independence graphs.
\newblock In {\em NeurIPS 2022 Workshop: New Frontiers in Graph Learning}, 2022.

\bibitem{shrivastava2022uglad}
Harsh Shrivastava, Urszula Chajewska, Robin Abraham, and Xinshi Chen.
\newblock \texttt{uGLAD}: Sparse graph recovery by optimizing deep unrolled networks.
\newblock {\em arXiv preprint arXiv:2205.11610}, 2022.

\bibitem{shrivastava2019glad}
Harsh Shrivastava, Xinshi Chen, Binghong Chen, Guanghui Lan, Srinvas Aluru, Han Liu, and Le~Song.
\newblock \texttt{GLAD}: Learning sparse graph recovery.
\newblock {\em arXiv preprint arXiv:1906.00271}, 2019.

\bibitem{shrivastava2015classification}
Harsh Shrivastava, Vijay Huddar, Sakyajit Bhattacharya, and Vaibhav Rajan.
\newblock Classification with imbalance: A similarity-based method for predicting respiratory failure.
\newblock In {\em 2015 IEEE international conference on bioinformatics and biomedicine (BIBM)}, pages 707--714. IEEE, 2015.

\bibitem{shrivastava2021system}
Harsh Shrivastava, Vijay Huddar, Sakyajit Bhattacharya, and Vaibhav Rajan.
\newblock System and method for predicting health condition of a patient, August~10 2021.
\newblock US Patent 11,087,879.

\bibitem{shrivastava2020grnular}
Harsh Shrivastava, Xiuwei Zhang, Srinivas Aluru, and Le~Song.
\newblock {GRNUlar}: Gene regulatory network reconstruction using unrolled algorithm from single cell {RNA}-sequencing data.
\newblock {\em bioRxiv}, 2020.

\bibitem{shrivastava2022grnular}
Harsh Shrivastava, Xiuwei Zhang, Le~Song, and Srinivas Aluru.
\newblock {GRNUlar}: A deep learning framework for recovering single-cell gene regulatory networks.
\newblock {\em Journal of Computational Biology}, 29(1):27--44, 2022.

\bibitem{sun2014monthly}
Alexander~Y Sun, Dingbao Wang, and Xianli Xu.
\newblock Monthly streamflow forecasting using gaussian process regression.
\newblock {\em Journal of Hydrology}, 511:72--81, 2014.

\bibitem{wang2019exact}
Ke~Alexander Wang, Geoff Pleiss, Jacob~R. Gardner, Stephen Tyree, Kilian~Q. Weinberger, and Andrew~Gordon Wilson.
\newblock Exact gaussian processes on a million data points.
\newblock In {\em arXiv}, 2019.

\bibitem{welling2011bayesian}
Max Welling and Yee~W Teh.
\newblock Bayesian learning via stochastic gradient langevin dynamics.
\newblock In {\em Proceedings of the 28th international conference on machine learning (ICML-11)}, pages 681--688, 2011.

\bibitem{wilkinson2019gaussian}
W~Wilkinson.
\newblock {\em Gaussian Process Modelling for Audio Signals}.
\newblock PhD thesis, Queen Mary University of London, 2019.

\bibitem{wilson2015thoughts}
Andrew~Gordon Wilson, Christoph Dann, and Hannes Nickisch.
\newblock Thoughts on massively scalable gaussian processes.
\newblock In {\em arXiv}, 2015.

\bibitem{wilson2016stochastic}
Andrew~Gordon Wilson, Zhiting Hu, Ruslan Salakhutdinov, and Eric~P. Xing.
\newblock Stochastic variational deep kernel learning.
\newblock In {\em arXiv}, 2016.

\bibitem{wilson2015kernel}
Andrew~Gordon Wilson and Hannes Nickisch.
\newblock Kernel interpolation for scalable structured gaussian processes (kiss-gp).
\newblock In {\em arXiv}, 2015.

\bibitem{yuan2019diverse}
Ye~Yuan and Kris Kitani.
\newblock Diverse trajectory forecasting with determinantal point processes.
\newblock In {\em arXiv}, 2019.

\bibitem{zheng2018dags}
Xun Zheng, Bryon Aragam, Pradeep~K Ravikumar, and Eric~P Xing.
\newblock {DAGs} with {NO TEARS}: Continuous optimization for structure learning.
\newblock {\em Advances in Neural Information Processing Systems}, 31:9472--9483, 2018.

\bibitem{zheng2020learning}
Xun Zheng, Chen Dan, Bryon Aragam, Pradeep Ravikumar, and Eric Xing.
\newblock Learning sparse nonparametric {DAGs}.
\newblock In {\em International Conference on Artificial Intelligence and Statistics}, pages 3414--3425. PMLR, 2020.

\end{thebibliography}
\bibliographystyle{plain}

\end{document}